\documentclass[runningheads]{llncs}
\usepackage{graphicx}
\usepackage{comment}
\usepackage{amsmath,amssymb} % define this before the line numbering.
\usepackage{color}
\usepackage{booktabs}
\begin{document}
% \linenumbers
\pagestyle{headings}
\mainmatter
\def\ECCVSubNumber{6055}  % Insert your submission number here

\title{Interpretable Foreground Object Search As Knowledge Distillation} % Replace with your title

%% INITIAL SUBMISSION 
%%\begin{comment}
%\titlerunning{ECCV-20 submission ID \ECCVSubNumber} 
%\authorrunning{ECCV-20 submission ID \ECCVSubNumber} 
%\author{Anonymous ECCV submission}
%\institute{Alibaba Group}
%%\end{comment}

\newcommand*\samethanks[1][\value{footnote}]{\footnotemark[#1]}

%******************

%\begin{comment}
\titlerunning{Interpretable Foreground Object Search As Knowledge Distillation}
\author{Boren Li\inst{1,}\thanks{Corresponding author. This work was partially supported by the National Key Research and Development Program of China (No.2018YFB1005002).} \and
Po-Yu Zhuang\inst{1} \and
Jian Gu\inst{1} \and Mingyang Li\inst{1} \and Ping Tan\inst{1,2}}
\authorrunning{Li et al.}
\institute{Alibaba Group \and Simon Fraser University\\
\email{\{boren.lbr, po-yu.zby, gujian.gj\}@alibaba-inc.com}, \email{mingyangli009@gmail.com}, \email{pingtan@sfu.ca}}
%\end{comment}
%******************
\maketitle

\begin{abstract}
This paper proposes a knowledge distillation method for foreground object search (FoS). Given a background and a rectangle specifying the foreground location and scale, FoS retrieves compatible foregrounds in a certain category for later image composition. Foregrounds within the same category can be grouped into a small number of patterns. Instances within each pattern are compatible with \textit{any} query input interchangeably. These instances are referred to as \textit{interchangeable foregrounds}. We first present a pipeline to build pattern-level FoS dataset containing labels of interchangeable foregrounds. We then establish a benchmark dataset for further training and testing following the pipeline. As for the proposed method, we first train a foreground encoder to learn representations of interchangeable foregrounds. We then train a query encoder to learn query-foreground compatibility following a knowledge distillation framework. It aims to transfer knowledge from interchangeable foregrounds to supervise representation learning of compatibility. The query feature representation is projected to the same latent space as interchangeable foregrounds, enabling very efficient and interpretable instance-level search. Furthermore, pattern-level search is feasible to retrieve more controllable, reasonable and diverse foregrounds. The proposed method outperforms the previous state-of-the-art by $10.42\%$ in absolute difference and $24.06\%$ in relative improvement evaluated by mean average precision (mAP). Extensive experimental results also demonstrate its efficacy from various aspects. The benchmark dataset and code will be release shortly.
\end{abstract}

\section{Introduction}\label{sec:intro}
Foreground object search (FoS) retrieves compatible foregrounds in a certain category given a background and a rectangle as query input \cite{Zhao_ECCV_2018}. It is a core task in many image composition applications \cite{Tsai_CVPR_2017}. For object insertion in photo editing, users often find it challenging and time-consuming to acquire compatible foregrounds in a foreground pool. Object insertion can be used to fill a new foreground to a region comprising undesired objects in the background \cite{Zhao_ICCV_2019}.

In a larger sense, for text-to-image synthesis with multiple objects, recent researches \cite{Johnson_2018_CVPR}\cite{Li_ICCV_2019} have shown insight to generate semantic layout at first. Then, one way to solve the follow-up task, layout to image, is multi-object retrieval and composition \cite{Chen_TOG_2009}. Directly retrieving multiple objects simultaneously suffers from combinatorial explosion that can be perfectly avoided by iteratively performing FoS with composition. Hence, FoS is also a significant underlying task.

\begin{figure}
	\begin{center}
		\includegraphics[width=10cm]{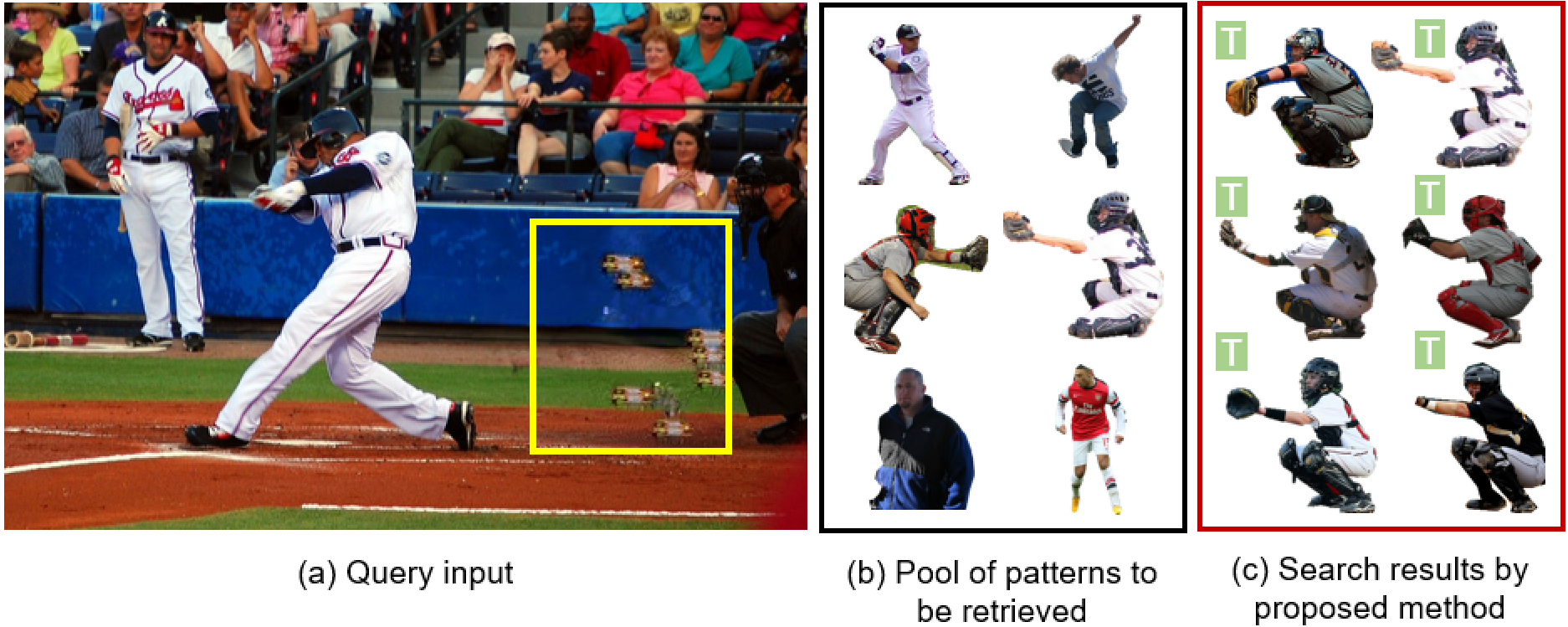}
	\end{center}
	\caption{Foreground object search (FoS). Given a background and a rectangle specifying the foreground location and scale as query input, FoS is to find compatible foregrounds within a certain category. (a) illustrates the query input. (b) exemplifies patterns in the foreground pool. (c) demonstrates search results by the proposed method.}
	\label{fig:motivation}
\end{figure}

Two problems arise to solve FoS. The first problem is how to classify foreground instances and define what are similar foregrounds to be retrieved together. The second problem is that given a query input and a foreground instance, how to define and decide their compatibility. Most recent methods \cite{Zhao_ECCV_2018}\cite{Zhao_ICCV_2019} jointly learned foreground similarity and query-foreground compatibility without decoupling the two problems. It makes the results difficult to interpret.

We notice that foregrounds in a certain category can be grouped to a small number of \textit{patterns}. Instances within the same pattern are compatible with \textit{any} query input interchangeably. These instances are referred to as \textit{interchangeable foregrounds}. Then, the first question arises: how to define and label interchangeable foregrounds specifically? 

Suppose we have answered the first question well, manually labelling compatibility for many pairs of query-foreground data is still extremely challenging, if not impossible. Since definition of interchangeable foregrounds relates to compatibility, the second question is: can we transfer knowledge from labelled interchangeable foregrounds to supervise representation learning of compatibility?

We answer these two questions in this work. For the first question, we propose a pipeline to build pattern-level FoS dataset comprising labels of interchangeable foregrounds. We exemplify `person' as the foreground category to explain how to label and establish a benchmark dataset for further training and testing. We then train a foreground encoder to classify these patterns in order to learn feature representations for interchangeable foregrounds. 

For the second question, we train a query encoder to learn query-foreground compatibility. It learns to transform query inputs into query features such that the feature similarities between query and compatible foregrounds are closer than those between query and incompatible ones. We follow a knowledge distillation scheme to transfer interchangeable foregrounds labelling to supervise compatibility learning. More specifically, we freeze the trained foreground encoder as the teacher network to generate embeddings as `soft targets' to train the query encoder in the student network. As a result, the query inputs are projected to the same latent space as interchangeable foregrounds, enabling very efficient and interpretable instance-level search. Furthermore, as interchangeable foregrounds are grouped into patterns, pattern-level search is feasible to retrieve more controllable, reasonable and diverse foregrounds.

We first show effectiveness of the foreground encoder to represent interchangeable foregrounds. We then demonstrate efficacy of the query encoder to represent query-foreground compatibility. The proposed method outperforms the previous state-of-the-art by $10.42\%$ in absolute difference and $24.06\%$ in relative improvement evaluated by mean average precision (mAP).

The key contributions are summarized as follows:
\begin{itemize}
	\item We introduce a novel concept called interchangeable foregrounds. It allows interpretable and direct learning of foreground similarity specifically for FoS. In addition, it makes pattern-level search feasible to retrieve more controllable, reasonable and diverse foregrounds.
	\item We propose a new pipeline to establish pattern-level FoS dataset containing labels of interchangeable foregrounds. We establish the first benchmarking dataset using this pipeline. This dataset will be released to the public.
	\item We propose a novel knowledge distillation framework to solve FoS. It enables fully interpretable learning and outperforms the previous state-of-the-art by a significant margin.
\end{itemize}

\section{Related Works}\label{sec:related_works}
\subsection{Foreground Object Search}
Early efforts on FoS, such as Photo Clip Art \cite{Laonde_TOG_2007} and Sketch2Photo \cite{Chen_TOG_2009}, applied handcrafted features to search foregrounds according to matching criterion as camera orientation, lighting, resolution, local context and so on. Manually designing either these matching criterion or handcrafted features is challenging. With the success of deep learning on image classification \cite{Russakovsky_IJCV_2015}, deep features are involved to replace handcrafted ones. Tan et al. \cite{Tan_WACV_2017} employed local region retrieval using semantic features extracted from an off-the-shelf CNN model. The retrieved regions contain person segments which are further used for image composition. They assume the foregrounds have surrounding background context and therefore, not feasible when the foregrounds are just images with pure background. Zhu et al. \cite{Zhu_ICCV_2015} trained a discriminative network to decide the realism of a composite image. They couple the suitability of foreground selection, adjustment and composition into one realism score, making it difficult to interpret.

Zhao et al. \cite{Zhao_ECCV_2018} first formally defined the FoS task and focused on the foreground selection problem alone. They applied end-to-end feature learning to adapt for different object categories. This work is the closest to ours and serves as the baseline method for comparison purpose. More recently, Zhao et al. \cite{Zhao_ICCV_2019} proposed an unconstrained FoS task that aims to retrieve universal compatible foreground without specifying its category. We only focus on the constrained FoS problem with known foreground category in this work.

\subsection{Knowledge Distillation}
Knowledge distillation is a general purpose technique that is widely applied for neural network compression \cite{Mishra_ICLR_2018}. The key idea is to use soft probabilities of a larger teacher network to supervise a smaller student network, in addition to the available class labels. The soft probabilities reveal more information than the class labels alone that can purportedly help the student network learn better.

In addition to neural network compression, prior works has found knowledge distillation to be useful for sequence modelling \cite{Kim_EMNLP_2016}, domain adaptation \cite{Meng_ICASSP_2018}, semi-supervised learning \cite{Tarvainen_NIPS_2017} and so on. The closest work to ours is multi-modal learning \cite{Gupta_CVPR_2016}. They trained a CNN model for a depth map as a new modality by teaching the network to reproduce the mid-level semantic representations learned from a well-labelled RGB image with paired data. For our case, we learn query-foreground compatibility as a new modality by teaching the network to reproduce the mid-level foreground similarity representations learned from a well-labelled interchangeable foreground modality with paired data. Therefore, the proposed FoS method can be viewed as another knowledge distillation application.

\section{Foreground Object Search Dataset}\label{sec:dataset}
In this section, we describe the proposed pipeline to build pattern-level FoS dataset containing labels of interchangeable foregrounds. We exemplify `person' as the foreground category to explain how to label and establish a benchmark dataset for further training and testing. Building a benchmark dataset is necessary for two reasons. First, there is no publicly available dataset for FoS. We do not have access to the one established by the baseline method \cite{Zhao_ECCV_2018}. Second, the previous dataset is instance-level and not sufficient to validate our method.

\begin{figure}
	\begin{center}
		\includegraphics[width=10cm]{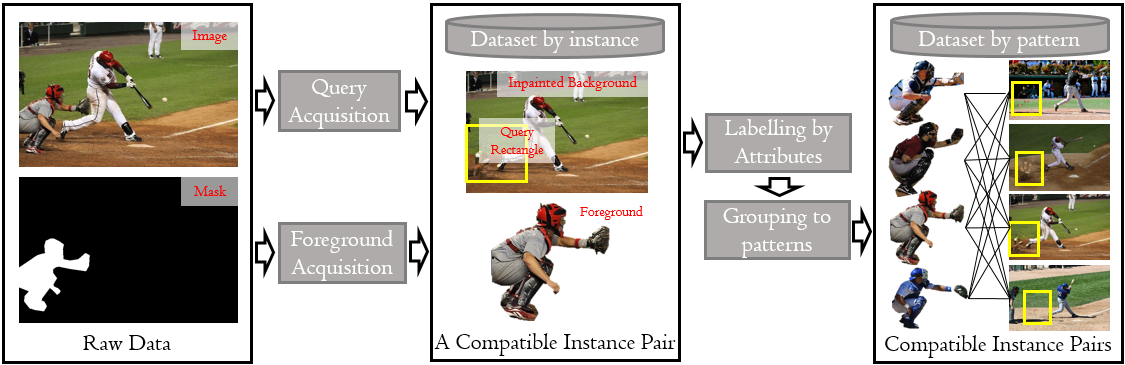}
	\end{center}
	\caption{The proposed pipeline to establish a pattern-level FoS dataset. The instance-level dataset is first established to obtain compatible instance pairs transformed from instance segmentation annotations. Through grouping interchangeable foregrounds to patterns, pattern-level dataset is finally built. A compatible instance pair in the instance-level dataset can be augmented to many pairs in the pattern-level dataset.}
	\label{fig:dataset_acquisition}
\end{figure}

\subsection{Pipeline to Establish Pattern-level FoS Dataset}
Fig. \ref{fig:dataset_acquisition} demonstrates the general pipeline to establish pattern-level FoS dataset. There exists publicly available datasets that contain instance segmentation masks, such as MS-COCO \cite{Lin_ECCV_2014}, PASCAL VOC 2012 \cite{Lin_ECCV_2014} and ADE20K \cite{Zhou_CVPR_2017}. We can decompose an image into a background scene, a foreground and a rectangle using a mask. Since they are all from the same image, they are naturally compatible. 

Previous methods \cite{Zhao_ECCV_2018}\cite{Zhao_ICCV_2019} leave the original foreground in the background scene when building the dataset. They do so because they mask out the foreground by a rectangle filled with image mean values during training with an early-fusion strategy. By contrast, we apply a free-form image inpainting algorithm \cite{Yu_ICCV_2019} to fill the foreground region in the background scene when building the dataset. This is because the deep inpainting algorithm trained on millions of images can perform reasonably well on this task. On the other hand, the early-fusion strategy by previous methods masks out too much background context, leaving the compatibility decision much more difficult. As for foreground samples in the dataset, we paste the foreground in the original image to the center location on a pure white square background.

With sufficient number of foregrounds in a certain category, the next goal is to group them into patterns of interchangeable foregrounds. Given many thousands of instances, this task is very challenging without supervision. Hence, we label foregrounds by attributes at first. We then group them into the same pattern if they have identical values in every attribute dimension. Finally, we establish a pattern-level dataset where much more compatible instance pairs can be extracted than its instance-level counterpart.

\subsection{Interchangeable Foregrounds Labelling}
We show how to label interchangeable foregrounds by using `person' as the foreground category. `person' is adopted because it is one of the most frequent categories for image composition. Furthermore, it is a non-rigid object with numerous different states. It is sufficiently representative to address the issues for interchangeable foregrounds labelling. We do not consider style issues in this work since all the raw images are photographs.

\begin{figure}
	\begin{center}
		\includegraphics[width=10cm]{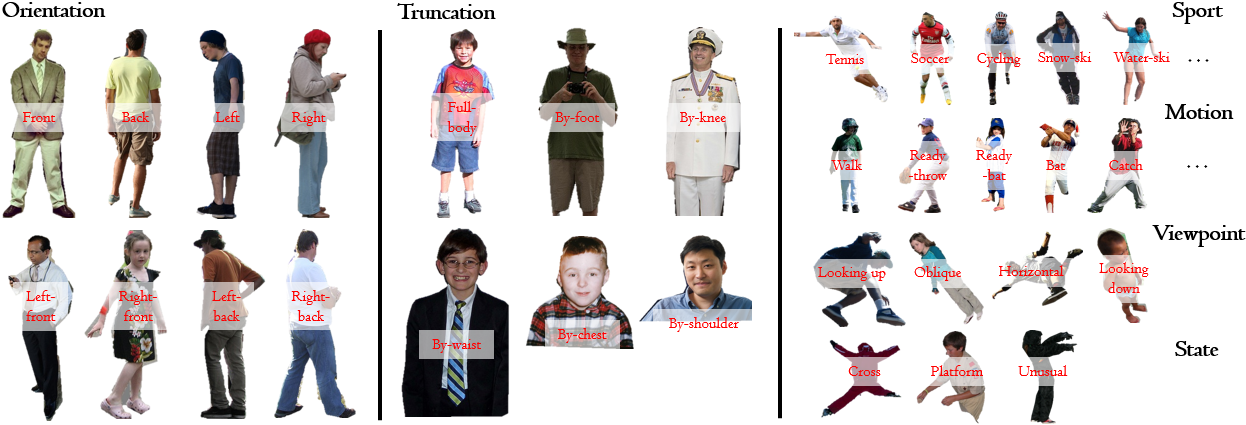}
	\end{center}
	\caption{An illustration of attributes for the `person' foreground. It contains six attribute dimensions: orientation, truncation, sport, motion, viewpoint and state. For a particular foreground, orientation and truncation are mandatory dimensions to be assigned with the presented values while the others are optional.}
	\label{fig:dataset}
\end{figure}

Fig. \ref{fig:dataset} illustrates the six attribute dimensions we defined to classify patterns of interchangeable foregrounds. For a particular foreground, \textit{orientation} and \textit{truncation} are two mandatory attribute dimensions to be assigned with the presented values. They are mandatory because they will largely determine most aspects of interchangeable foregrounds. The other four attribute dimensions are sport, motion, viewpoint and state. These dimensions can further distinguish various aspects of `person'. Their values can be left as `unspecified' when we cannot assign them with available values. Table \ref{tab:attr_val_each_dim} shows the number of available attribute values in each dimension. 

\begin{table}
	\caption{Number of available attribute values in each dimension}
	\begin{center}
		\begin{tabular}{c|c|c|c|c|c}
			\toprule[1.2pt]
			Orientation & Truncation & Sport & Motion & Viewpoint & State\\
			\hline\hline
			$8$ & $6$ & $12$ & $31$ & $4$ & $3$\\
		\bottomrule[1.2pt]
		\end{tabular}
	\end{center}
%	\caption{Number of available attribute values in each dimension}
	\label{tab:attr_val_each_dim}
\end{table}

We adopt images with mask annotations in the MS-COCO \cite{Lin_ECCV_2014} dataset as raw data. Before labelling attribute values for each sample, we first exclude inappropriate samples that are heavily occluded, small or incomplete, resulting in $10154$ foregrounds. We label $5468$ samples from them with these attribute values, leading to $699$ different patterns after grouping. Thus, we obtain $5468$ pattern-level query-foreground compatibility pairs in total. Furthermore, the remaining $4686$ unannotated foregrounds can be labelled automatically by a trained foreground encoder presented in Section \ref{sec:proposed_approach}. It leads to more pairs of pattern-level data to train query-foreground compatibility. In a larger sense, applying our trained foreground encoder with an instance segmentation model such as Mask-RCNN \cite{He_ICCV_2017}, we can automate the whole pipeline using internet images to learn query-foreground compatibility.

\subsection{Evaluation Set and Metrics}
The annotated foreground patterns follow a heavy-tailed distribution. Therefore, we only select those patterns with at least $20$ interchangeable foregrounds for testing. This leads to $69$ patterns in total. We randomly select $5$ foreground instances from each of these patterns to obtain the foreground database at test time. These foregrounds can be also applied to evaluate the capability of the foreground encoder in classifying interchangeable foregrounds. We adopt top-$1$ and top-$5$ accuracies to evaluate the classifier with $699$ classes altogether.

Simultaneously, we obtain the same number of corresponding query inputs. We select $100$ query samples and prefer those with more `person' in the query background intentionally to make the dataset more challenging. We then manually label their compatibility to each foreground pattern in the test-time foreground database. This is because one query input may have multiple other compatible foreground patterns except the corresponding one. On average, for each query input, we label $22.35$ and $6.07$ compatible foreground instances and patterns, respectively.
These pairs are employed to evaluate query-foreground compatibility. We adopt mAP to evaluate the overall performance of FoS.

\section{Proposed Approach}\label{sec:proposed_approach}

\subsection{Overall Training Scheme}\label{subsec:overall_train_scheme}
\begin{figure}
	\begin{center}
		\includegraphics[width=10cm]{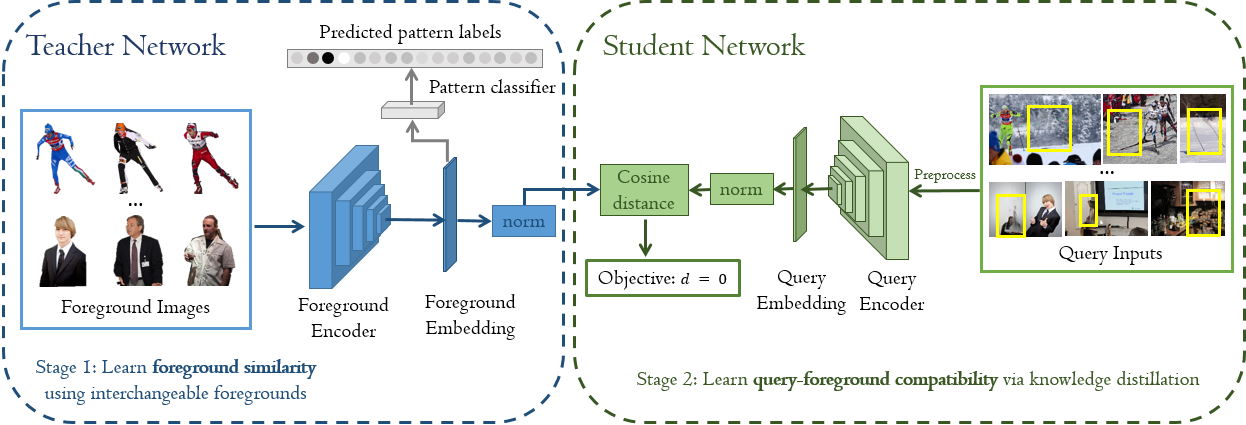}
	\end{center}
	\caption{The overall training scheme. The foreground encoder is first trained to classify patterns of interchangeable foregrounds. It then serves as the teacher network to generate foreground embeddings as soft targets to train the query encoder which encodes query-foreground compatibility. }
	\label{fig:overall_train_scheme}
\end{figure}

Fig. \ref{fig:overall_train_scheme} presents the overall training scheme comprising two successive stages. The first stage trains the foreground encoder to classify patterns of interchangeable foregrounds in order to learn foreground feature representations. Feature similarities from the same pattern are closer than those from other patterns. Therefore, the learned features are fully interpretable.

The second stage trains the query encoder to learn \textit{query-foreground compatibility}. This encoder transforms query inputs into embeddings such that embedding distances between query and compatible foregrounds are closer. We aim to transfer the knowledge of interchangeable foregrounds labelling to supervise compatibility learning. Hence, during training, we freeze the foreground encoder trained from the first stage as the teacher network. It generates foreground embeddings as `soft targets' to train the query encoder in the student network. As a result, the query inputs are projected to the same latent space as interchangeable foregrounds, enabling very efficient and interpretable instance-level search. Cosine distance is applied to measure embedding distances between query and foreground. The embeddings are $l_2$ normalized before computing cosine distance.

\subsection{Foreground Encoder}
Training for the foreground encoder follows a typical image classification pipeline. The deeply learned embeddings need to be not only \textit{separable} but also \textit{discriminative}. These embeddings require to be well-classified by k-nearest neighbour algorithms without necessarily depend on label prediction.

Therefore, we adopt center loss \cite{Wen_ECCV_2016} in addition to softmax loss to train more discriminative features. The center loss is used due to its proven success in the face recognition task that is very similar to ours. The loss function is given by
\begin{equation}\label{eq:fg_loss}
\mathcal{L}^f = \mathcal{L}^f_S + \lambda\mathcal{L}^f_C\text{.}
\end{equation}
$\mathcal{L}^f$ denotes the total loss for foreground classification. The superscript $f$ denotes \textit{foreground} later on. $\mathcal{L}^f_S$ is the conventional softmax loss. $\mathcal{L}^f_C$ is the center loss and $\lambda$ is the weight. $\mathcal{L}^f_C$ is given by
\begin{equation}\label{eq:fg_center_loss}
\mathcal{L}^f_C = \frac{1}{2}\sum_{i=1}^{m}\|\mathbf{x}_i^f - \mathbf{c}_{y_i}^f\|_2^2\text{,}
\end{equation}
where $m$ is the batch size, $\mathbf{x}_i^f\in\mathbb{R}^d$ denotes the $i^{th}$ embedding, and $\mathbf{c}_{y_i}^f\in\mathbb{R}^d$ is the embedding center of the $y_i^{th}$ pattern. $d$ is the feature dimension.

As for the foreground encoder architecture, we adopt ResNet50 \cite{He_CVPR_2016} with $2048$ dimensional feature embedding as feature extractor. We initialize the weights that were pre-trained for the ILSVRC-2014 competition \cite{Russakovsky_IJCV_2015}. A fully connected layer is further appended to the feature extractor for pattern classification.

\subsection{Query Encoder}
Compatibility is determined by three factors: the background context, the foreground context, and the foreground location and scale (i.e. layout). We do not consider style compatibility in this work, but our framework is fully adaptable to style encodings learned from \cite{Collomosse_ICCV_2017}. We focus to retrieve compatible foregrounds in a certain category without considering the multi-class problem, since our work can be easily expanded using \cite{Zhao_ECCV_2018} to tackle this issue. It is still challenging to hand-design compatibility criterion, even considering only the three factors.
\begin{figure}
	\begin{center}
		\includegraphics[width=10cm]{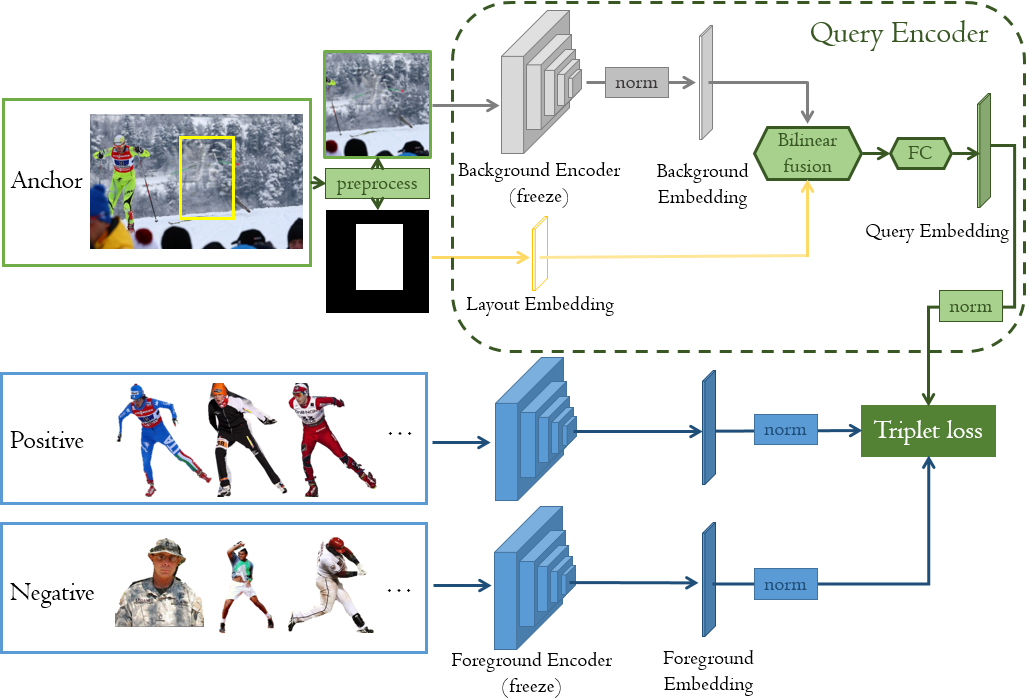}
	\end{center}
	\caption{The training scheme for the query encoder as knowledge distillation. The query encoder is trained to generate query embeddings using a triplet network. This network aims to distill compatibility information from the foreground encoder trained to represent patterns of interchangeable foregrounds.}
	\label{fig:query_encoder}
\end{figure}

\subsubsection{Network Architecture}
Fig. \ref{fig:query_encoder} demonstrates the training scheme for the query encoder as knowledge distillation. This encoder transforms query inputs into embeddings such that embedding distances between query and compatible foregrounds are closer. The general architecture follows a typical two-stream network. The bottom stream takes the square foreground image with pure white background as input. It encodes the image to feature embedding using the foreground encoder trained in the first stage. We freeze the weights in the foreground encoder during training for the query encoder.

The top stream takes a background scene and a rectangle specifying the desired foreground location and scale as query input. The background scene is first cropped to a square image, where the desired foreground location is placed as close to the image center as possible. This cropping also preserves as much context as possible for the square-background. Such cropping makes the background image more consistent so that the training is more stable. The square-background is encoded by a ResNet50 \cite{He_CVPR_2016} backbone pre-trained on ImageNet \cite{Russakovsky_IJCV_2015} with $2048$-dimensional features. This network serves as the background encoder to represent scene context. Since the pre-trained network can represent semantic context well, we freeze its weights during training for the query encoder. 

The query rectangle is just a bounding box with four degrees of freedom (DoF). We adopt the centroid representation for the bounding box. The first two DoF are coordinates of the bounding box centroid. The other two DoF are width and height of the bounding box. These coordinates are then normalized by dividing the image side length. We only keep the first two digits after the decimal point for better generalization of the bounding box encoding. This encoding is referred to as \textit{layout embedding}. 

Unlike previous methods \cite{Zhao_ECCV_2018}\cite{Zhao_ICCV_2019} by filling the query rectangle with image mean values to the background scene as a unified query input, our method encodes the two query factors separately to make the embeddings more interpretable. In addition, previous methods may fail when the query rectangle is too big relative to the background scene because too few background context can be preserved after the early-fusion to a unified query input. By contrast, we can avoid this issue completely since we encode the full square-background context. This is feasible because the foreground object has already been removed from the background scene when we establish the FoS dataset.

The layout and background embeddings are late-fused using bilinear fusion \cite{Lin_ICCV_2015}. Here, the two embeddings are fused using their outer product followed by flattening to a vector. The outer product is adopted since it can model pairwise feature interactions well. Because the layout embedding is only 4-dimensional, we have not applied compact bilinear pooling techniques \cite{Gao_CVPR_2016}\cite{Fukui_EMNLP_2016}\cite{Yu_ICCV_2017} to reduce the dimension of the fused feature. This feature is then transformed by two fully connected (FC) layers with ReLU activation to obtain the query embedding. The output dimensions for the first and second FC are all $2048$.

\subsubsection{Loss Function}
We construct triplets consisting of a query input as anchor, a compatible foreground as positive, and an incompatible foreground as negative to train the network. We adopt triplet loss \cite{Schroff_CVPR_2015} and enforce the embedding distance between anchor and positive to be closer than the one between anchor and negative. These embeddings are $l_2$ normalized before measuring distance using cosine function. 

Formally,  a fused feature after bilinear fusion is given by $\mathbf{u}^q\in\mathbb{R}^e$, where the superscript $q$ denotes \textit{query} later on and $e$ is the dimension of the feature embedding. Denote the foreground embeddings for the positive and negative samples are $\mathbf{x}_p^f$ and $\mathbf{x}_n^f$, respectively. The operation of two FC layers with ReLU is denoted as $\mathcal{F}$. The triplet loss is then given by
\begin{equation}\label{eq:triplet_loss}
\mathcal{L}^q = \max\left(0, \frac{\mathcal{F}\left(\mathbf{u}^q\right)^T\mathbf{x}_p^f}{\|\mathcal{F}\left(\mathbf{u}^q\right)\|\|\mathbf{x}_p^f\|} - \frac{\mathcal{F}\left(\mathbf{u}^q\right)^T\mathbf{x}_n^f}{\|\mathcal{F}\left(\mathbf{u}^q\right)\|\|\mathbf{x}_n^f\|} + M\right),
\end{equation}
where $M$ is a positive margin. The objective is to train $\mathcal{F}$ by minimizing $\mathcal{L}^q$ over all the sampled triplets.

\subsubsection{Training Data}
The pattern-level FoS dataset is used for training. The dataset contains pairs of query and compatible pattern containing interchangeable foreground instances. A query with these instances form positive pairs, whereas the query with the others are all negative ones. With pattern-level FoS dataset, we can largely alleviate the severe imbalance in the number of training samples, coupled with noise in the negative pair sampling where some compatible foregrounds are mistreated as negative ones.

We apply different data augmentation strategies for the three types of input. To augment the query rectangle, we relax its size and scale constraints by randomly resizing the rectangle with maximum possible space being half of the rectangle's width and height. To augment the query background, we add random zoom on the cropped square-background while keep the whole query rectangle within the field of view. This augmentation strategy cannot be applied by previous methods \cite{Zhao_ECCV_2018}\cite{Zhao_ICCV_2019} since it will result in fewer background context in the early-fused query input. As for foreground augmentation, we adopt the same strategy when training the foreground encoder.

\subsection{Pattern-level Foreground Object Search}
With the novel concept of interchangeable foregrounds, we can apply pattern-level FoS instead of instance-level. For each foreground instance in the query database, we can assign a pattern label on it. Having all foreground instances within a pattern, the pattern embedding is computed using the centroid of all the instance embeddings transformed by the trained foreground encoder. These pattern embeddings can be also indexed for retrieval. Pattern-level FoS can easily stratify the results, making it more feasible to retrieve controllable, reasonable and diverse foreground instances.

\subsection{Implementation Details}
To train the foreground encoder, we use the SGD optimizer with momentum and weight decay set to $0.9$ and $0.0001$, respectively. The learning rate for the softmax loss is $0.02$ and the learning rate decay is $0.5$ for every $10$ epochs. The center loss weight, $\lambda$, is set to $0.005$. The learning rate for the center loss is $0.5$. Batch size is $32$ during training. For offline augmentation, we add random padding to the foreground and fill in the padded region with white color. Each foreground is augmented to $20$ samples. We then pad them to square images with pure white background. For online augmentation, we apply color jitter by randomly changing the brightness, saturation, contrast and hue by $0.4$, $0.4$, $0.4$ and $0.2$, respectively. These samples are resized to $256\times256$ before fed into the foreground encoder.

To train the query encoder, we use the Adam optimizer \cite{Kingma_2014_CoRR} with $\beta_1=0.5$, $\beta_2=0.99$ and $\epsilon=10^{-9}$. The learning rate is $10^{-4}$ for the triplet loss. Batch size is $16$ during training. The margin, $M$, is set to $0.1$. The input size of the background encoder is $256\times256$. We perform offline augmentations as described. Each query-foreground pair is augmented to $20$ samples. For online augmentation, we apply color jitter by randomly changing the brightness, saturation, contrast and hue by $0.4$, $0.4$, $0.4$ and $0.2$, respectively.

\begin{figure}
	\begin{center}
		\includegraphics[width=10cm]{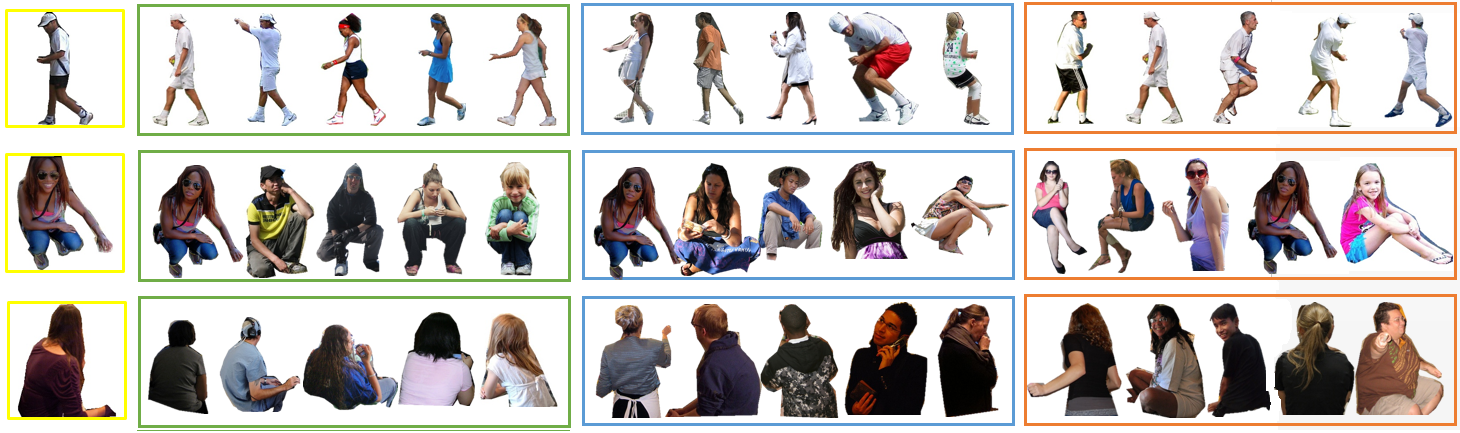}
	\end{center}
	\caption{Retrieval results comparison on similar foregrounds. The yellow box denotes the query foreground. The top-$5$ most similar foregrounds retrieved by the proposed method, the baseline method \cite{Zhao_ECCV_2018} and the pre-trained ResNet50 model are shown in the green, blue and orange boxes, respectively.}
	\label{fig:fg_similarity_comparison}
\end{figure}

\section{Experiments}\label{sec:exp}
\subsection{Foreground Encoder}
We train foreground encoder in the first stage to classify patterns of interchangeable foregrounds. We use a foreground as query and search for its top-$5$ most similar foregrounds in a large database comprising $10154$ samples. We first encode all the samples into embeddings using our trained foreground encoder. These embeddings are further $l_2$ normalized for query using cosine distance. We apply brute-force k-nearest neighbour matching to obtain the retrieval results. We compare results with the baseline method \cite{Zhao_ECCV_2018} and the pre-trained ResNet50 model on ImageNet as shown by Fig. \ref{fig:fg_similarity_comparison}. Clearly, similar instances retrieved by our method are much more interpretable. We can also apply pattern-level search to create interpretable and controllable diversity.

To further quantify the performance of foreground encoder as a pattern classifier, we test it on our evaluation set. The top-$1$ and top-$5$ accuracies are respectively $53.15\%$ and $85.79\%$ with $699$ classes. The accuracy can be further improved with more labelled data, while the trained foreground encoder is sufficient to achieve much better performance over the baseline method in supervising query-foreground compatibility later.

\subsection{Query Encoder}
We compare our results with the baseline method \cite{Zhao_ECCV_2018}. We remove the MCB module in the baseline method since we only focus on FoS with one foreground category. Since their implementation is not publicly available, we implement it by strictly following all the settings in their paper. We train both methods on the newly established FoS dataset. We prepare $2$ million triplets for each method and train for $2$ epochs until convergence. 

\begin{figure}
	\begin{center}
		\includegraphics[width=10cm]{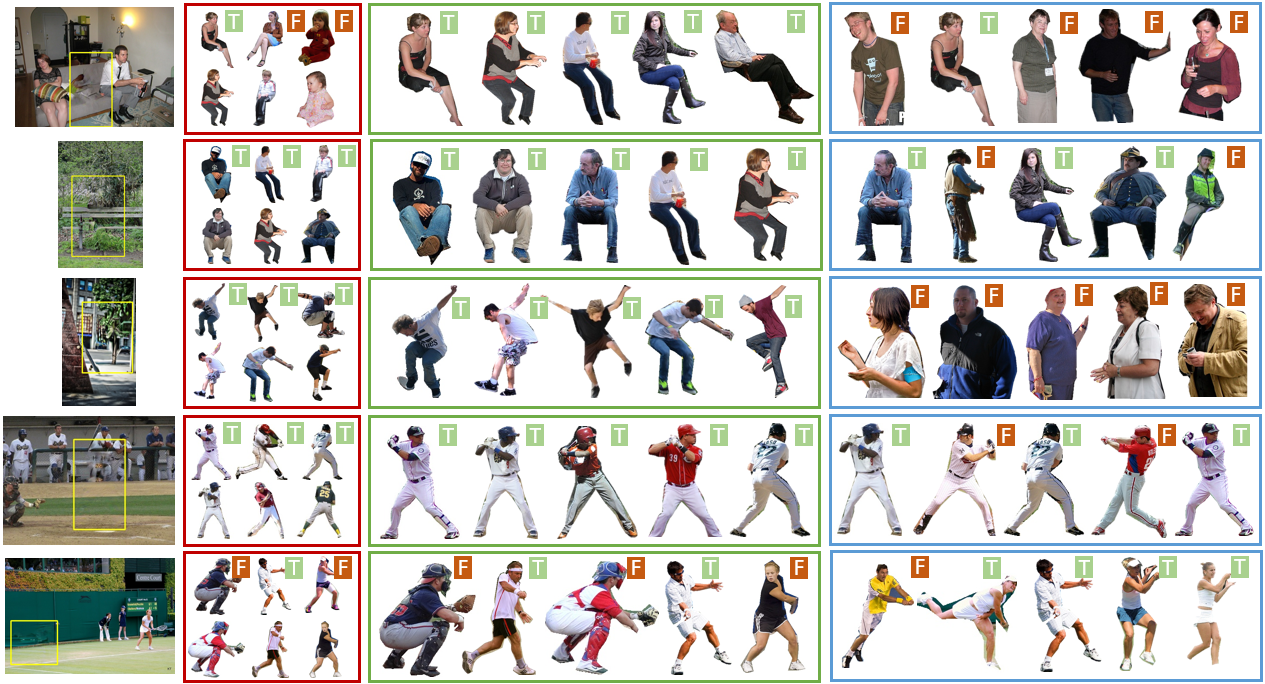}
	\end{center}
	\caption{Retrieval results comparison with baseline method \cite{Zhao_ECCV_2018}. Each row represents one query. The leftmost image demonstrates the query input. The red box shows top-$3$ patterns from pattern-level search, each with top-$2$ instances shown in a column, using the proposed method. The top-$5$ instance-level search results by our method and the baseline method are shown in the green and blue boxes, respectively.}
	\label{fig:comparison}
\end{figure}

We first compare results from the two methods qualitatively in Fig. \ref{fig:comparison}. Each row represents one query. The leftmost image shows the query input. Results from pattern- and instance-level search using our method are given in the red and green boxes, respectively. The instance-level search results from the baseline method are shown in the blue box. As can be seen, pattern-level search can provide reasonable and diverse results in a more controllable fashion than instance-level search. As for instance-level search, our results are much more reasonable and interpretable as seen from the first to third row. When the query rectangle is big relative to the background image, the baseline method cannot work properly due to its early-fusion strategy in the query stream. The third row illustrates such a case where a skateboard appears in the background image but most parts of the skateboard are within the query rectangle. The baseline method masks out this crucial cue with early-fusion, resulting in the fatal errors. Our method uses late-fusion without losing any information from the query inputs and therefore, it easily captures the important cue within the query rectangle. Results in the forth and fifth row demonstrate a limitation of both the proposed and baseline method. This limitation originates from the preprocessing step that square-crops the background image. Take the case in the fifth row for example. After square-cropping the query background, the woman playing tennis on the opposite side to the query rectangle is completely cropped, resulting in the final confusion of the retrieval results.

Quantitatively, we test both methods on our evaluation set. The mAP is $43.30\%$ using the baseline method whereas ours is $53.72\%$. It outperforms the baseline by $10.42\%$ in absolute difference and $24.06\%$ in relative improvement.

\subsubsection{Ablation Study}
\begin{table}
	\caption{Ablation study results with mAP in percentage. `Baseline' denotes baseline method \cite{Zhao_ECCV_2018}. `Early-fusion' denotes training using early-fused query inputs. `No-aug' denotes late fusion without random zoom augmentation. `No-bg-freeze' denotes training without freezing background context encoder. `Multi-task' denotes training using a multi-task loss to jointly train the foreground and query encoder.}
	\begin{center}
		\begin{tabular}{c|c|c|c|c|c}
			\toprule[1.5pt]
			Baseline & Early-fusion & No-aug  & No-bg-freeze & Multi-task & Ours\\
			\hline\hline
			$43.30$ & $48.98$ \color{blue}{5.68}$\uparrow$ & $51.91$ \color{blue}{8.61}$\uparrow$& $53.61$ \color{blue}{10.31}$\uparrow$ & $54.48$ \color{blue}{11.18}$\uparrow$& $\mathbf{53.72}$ \color{blue}{10.42}$\uparrow$\\
			\bottomrule[1.2pt]
		\end{tabular}
	\end{center}
	\label{tab:ablation}
\end{table}

Table \ref{tab:ablation} shows results in mAP of five ablation variants. The value in blue shows their respective absolute changes relative to the baseline method. We first investigate the significance to apply interchangeable foregrounds. We employ early fusion strategy in the query stream similar to the baseline method, while we keep our pre-training for interchangeable foregrounds. With the newly introduced interchangeable foregrounds pre-training, the mAP is enhanced by $5.68\%$, contributing to $54.51\%$ for the overall improvement. In the second variant, we apply our late fusion strategy in the query stream without random zoom augmentation. It further improves the mAP by $2.93\%$, contributing to $28.12\%$ for the overall improvement. In the third experiment, we add random zoom augmentation. The baseline method \cite{Zhao_ECCV_2018} cannot perform this augmentation since in many cases, the zoomed background with masked query rectangle lacks background context. In this experiment, we do not freeze the background encoder. With this augmentation, the mAP is further enhanced by $1.7\%$, contributing to $16.31\%$ for the overall improvement. In the fourth experiment, we freeze the background encoder and just train the two FC with ReLU layers. Results have shown that training for the background encoder simultaneously cannot help determining compatibility. It implies that the pre-trained model is sufficient to encode semantic context well for the background. In the final ablation experiment, we further fine-tune the foreground and query encoder with a multi-task loss without freezing the foreground encoder. It gives a gain of $0.76\%$. However, the gain will be less as we enlarge the interchangeable foreground dataset. By contrast, our knowledge distillation framework can modularize FoS into two sub-tasks whose dataset can be prepared separately.

\section{Conclusions}
This paper introduces a novel concept called interchangeable foregrounds for FoS. It enables interpretable and direct learning of foreground similarity. It also makes pattern-level search feasible to retrieve controllable, reasonable and diverse foregrounds. A new pipeline is proposed to build pattern-level FoS dataset with labelled interchangeable foregrounds. The first FoS benchmark dataset is established accordingly. A novel knowledge distillation framework is proposed to solve the FoS task. It provides fully interpretable results and enhances the absolute mAP by $10.42\%$ and relative mAP by $24.06\%$ over the previous state-of-the-art. It implies the knowledge from interchangeable foregrounds can be transferred to supervise compatibility learning for better performance.

\clearpage
\bibliographystyle{splncs04}
\bibliography{egbib}

\begin{thebibliography}{10}
\providecommand{\url}[1]{\texttt{#1}}
\providecommand{\urlprefix}{URL }
\providecommand{\doi}[1]{https://doi.org/#1}

\bibitem{Chen_TOG_2009}
Chen, T., Cheng, M.M., Tan, P., Shamir, A., Hu, S.M.: Sketch2photo: internet
  image montage. ACM Trans. on Graphics  (2009)

\bibitem{Collomosse_ICCV_2017}
Collomosse, J., Bui, T., Wilber, M., Fang, C., Jin, H.: Sketching with style:
  visual search with sketches and aesthetic context. In: ICCV (2017)

\bibitem{Fukui_EMNLP_2016}
Fukui, A., Park, D.H., Yang, D., Rohrbach, A., Darrell, T., Rohrbach, M.:
  Multimodal compact bilinear pooling for visual question answering and visual
  grounding. In: EMNLP (2016)

\bibitem{Gao_CVPR_2016}
Gao, Y., Beijbom, O., Zhang, N., Darrell, T.: Compact bilinear pooling. In:
  CVPR (2016)

\bibitem{Gupta_CVPR_2016}
Gupta, S., Hoffman, J., Malik, J.: Cross modal distillation for supervision
  transfer. In: CVPR (2016)

\bibitem{He_ICCV_2017}
He, K., Gkioxari, G., Dollár, P., Girshick, R.: Mask r-cnn. In: ICCV (2017)

\bibitem{He_CVPR_2016}
He, K., Zhang, X., Ren, S., Sun, J.: Deep residual learning for image
  recognition. In: CVPR (2016)

\bibitem{Johnson_2018_CVPR}
Johnson, J., Gupta, A., Fei-Fei, L.: Image generation from scene graphs. In:
  CVPR. pp. 1219--1228 (Jun 2018)

\bibitem{Kim_EMNLP_2016}
Kim, Y., Rush, A.M.: Sequence-level knowledge distillation. In: EMNLP (2016)

\bibitem{Kingma_2014_CoRR}
Kingma, D.P., Ba, J.: Adam: a method for stochastic optimization. CoRR
  \textbf{arXiv:1412.6980} (2014)

\bibitem{Laonde_TOG_2007}
Lalonde, J.F., Hoiem, D., Efros, A.A., Rother, C., Winn, J., Criminisi, A.:
  Photo clip art. ACM Trans. on Graphics (TOG)  (2007)

\bibitem{Li_ICCV_2019}
Li, B., Zhuang, B., Li, M., Gu, J.: Seq-sg2sl: inferring semantic layout from
  scene graph through sequence to sequence learning. In: ICCV (2019)

\bibitem{Lin_ECCV_2014}
Lin, T., Maire, M., Belongie, S.J., Bourdev, L.D., Girshick, R.B., Hays, J.,
  Perona, P., Ramanan, D., Doll{\'{a}}r, P., Zitnick, C.L.: Microsoft {COCO:}
  common objects in context. In: ECCV (2014)

\bibitem{Lin_ICCV_2015}
Lin, T.Y., RoyChowdhury, A., Maji, S.: Bilinear cnn models for fine-grained
  visual recognition. In: ICCV (2015)

\bibitem{Meng_ICASSP_2018}
Meng, Z., Li, J., Gong, Y., Juang, B.H.: Adversarial teacher-student learning
  for unsupervised domain adaptation. In: ICASSP (2018)

\bibitem{Mishra_ICLR_2018}
Mishra, A., Marr, D.: Apprentice: using knowledge distillation techniques to
  improve low-precision network accuracy. In: ICLR (2018)

\bibitem{Russakovsky_IJCV_2015}
Russakovsky, O., Deng, J., Su, H., Krause, J., Satheesh, S., Ma, S., Huang, Z.,
  Karpathy, A., Khosla, A., Bernstein, M., Berg, A.C., Fei-Fei, L.: Imagenet
  large scale visual recognition challenge. IJCV  (2015)

\bibitem{Schroff_CVPR_2015}
Schroff, F., Kalenichenko, D., Philbin, J.: Facenet: a unified embedding for
  face recognition and clustering. In: CVPR (2015)

\bibitem{Tan_WACV_2017}
Tan, F., Bernier, C., Cohen, B., Ordonez, V., Barnes, C.: Where and who?
  automatic semantic-aware person composition. In: WACV (2017)

\bibitem{Tarvainen_NIPS_2017}
Tarvainen, A., Valpola, H.: Mean teachers are better role models:
  weight-averaged consistency targets improve semi-supervised deep learning
  results. In: NIPS (2017)

\bibitem{Tsai_CVPR_2017}
Tsai, Y., Shen, X., Lin, Z., Sunkavalli, K., Lu, X., Yang, M.: Deep image
  harmonization. In: CVPR (2017)

\bibitem{Wen_ECCV_2016}
Wen, Y., Zhang, K., Li, Z., Qiao, Y.: A discriminative feature learning
  approach for deep face recognition. In: ECCV (2016)

\bibitem{Yu_ICCV_2019}
Yu, J., Lin, Z., Yang, J., Shen, X., Lu, X., Huang, T.S.: Free-form image
  inpainting with gated convolution. In: ICCV (2019)

\bibitem{Yu_ICCV_2017}
Yu, Z., Yu, J., Fan, J., Tao, D.: Multi-modal factorized bilinear pooling with
  co-attentionlearning for visual question answering. In: ICCV (2017)

\bibitem{Zhao_ECCV_2018}
Zhao, H., Shen, X., Lin, Z., Sunkavalli, K., Price, B., Jia, J.:
  Compositing-aware image search. In: ECCV (2018)

\bibitem{Zhao_ICCV_2019}
Zhao, Y., Price, B., Cohen, S., Gurari, D.: Unconstrained foreground object
  search. In: ICCV (2019)

\bibitem{Zhou_CVPR_2017}
Zhou, B., Zhao, H., Puig, X., Fidler, S., Barriuso, A., Torralba, A.: Scene
  parsing through ade20k dataset. In: CVPR (2017)

\bibitem{Zhu_ICCV_2015}
Zhu, J.Y., Krahenbuhl, P., Shechtman, E., Efros, A.A.: Learning a
  discriminative model for the perception of realism in composite images. In:
  ICCV (2015)

\end{thebibliography}
\end{document}